%% file: main.tex
\documentclass[conference,10pt,letter]{IEEEtran}
\IEEEoverridecommandlockouts

\usepackage{times}
\usepackage{graphicx}
\usepackage[cmex10]{amsmath}
\usepackage{amssymb}
\usepackage{cite}
\usepackage{url}
\usepackage[utf8]{inputenc}
\usepackage{booktabs}
\usepackage{algorithm}
\usepackage{algorithmic}
\usepackage{subcaption}
\usepackage{multirow}
\usepackage{xcolor}
\usepackage{xspace}

\def\BibTeX{{\rm B\kern-.05em{\sc i\kern-.025em b}\kern-.08em
    T\kern-.1667em\lower.7ex\hbox{E}\kern-.125emX}}

\newcommand{\eg}{\textit{e}.\textit{g}.\@\xspace}

\begin{document}
\title{Adaptive Adversarial Attack on Scene Text Recognition}
\author{
\IEEEauthorblockN{Xiaoyong Yuan\IEEEauthorrefmark{1}, Pan He\IEEEauthorrefmark{1}, Xiaolin Li\IEEEauthorrefmark{2}, Dapeng Wu\IEEEauthorrefmark{1}}
\IEEEauthorblockA{\IEEEauthorrefmark{1}NSF Center for Big Learning, University of Florida\\
\IEEEauthorrefmark{2}AI Institute, Tongdun Technology  \\
chbrian@ufl.edu, pan.he@ufl.edu, xiaolin.li@tongdun.net, dpwu@ieee.org
}
}

\maketitle

\begin{abstract}
Recent studies have shown that state-of-the-art deep learning models are vulnerable to the inputs with small perturbations (adversarial examples). We observe two critical obstacles in adversarial examples: (i) Recent adversarial attacks require manually tuning hyper-parameters and take a long time to construct an adversarial example, making it impractical to attack real-time systems; (ii) Most of the studies focus on non-sequential tasks, such as image classification, yet only a few consider sequential tasks. In this work, we propose an adaptive approach to speed up adversarial attacks, especially on sequential learning tasks. By leveraging the uncertainty of each task, we directly learn the adaptive multi-task weightings, without manually searching hyper-parameters. A unified architecture is developed and evaluated for both non-sequential tasks and sequential ones. To evaluate the effectiveness, we take the scene text recognition task as a case study. To our best knowledge, our proposed method is the first attempt to adversarial attack for scene text recognition. \textit{Adaptive Attack} achieves over 99.9\% success rate with $3\sim6\times$ speedup compared to state-of-the-art adversarial attacks. 
\end{abstract}


\section{Introduction}
\input{intro}




\section{Adaptive Adversarial Attack}
\input{adv.tex}

\section{Experiments and Analysis}
\input{analysis.tex}


\section{Conclusions}
\input{conclusion.tex}
{\small
\bibliographystyle{ieee}
\bibliography{others,adversarial,deep}
}
\end{document}

%% file: intro.tex
Recent studies~\cite{szegedy2013intriguing,goodfellow2014explaining,carlini2017towards,yuan2019adversarial} have shown that deep neural networks are vulnerable to \textit{adversarial examples}, by adding imperceptible perturbations on original images to fool a deep learning model. Adversarial examples have raised significant concerns since deep learning models are prevalent in many security-critical systems. 


\textbf{Accelerating Adversarial Attacks:} Attacking deep learning models is time-limited in many real-world systems (\eg, autonomous driving, face recognition). 
Many strong iterative optimization-based attacks~\cite{kurakin2016adversarial,carlini2017towards,athalye2018obfuscated} generate adversarial examples of high-quality that are hard to be defended~\cite{carlini2017adversarial,athalye2018obfuscated}, but these iterative methods usually take a longer time to find proper weights/hyper-parameters for optimization, which becomes an obstacle to attacking real-time systems. For instance, it takes about one hour to generate a piece of adversarial audio with a few seconds~\cite{carlini2018audio}.

\begin{figure}[!tb]
\centering
    \includegraphics[width=0.8\linewidth]{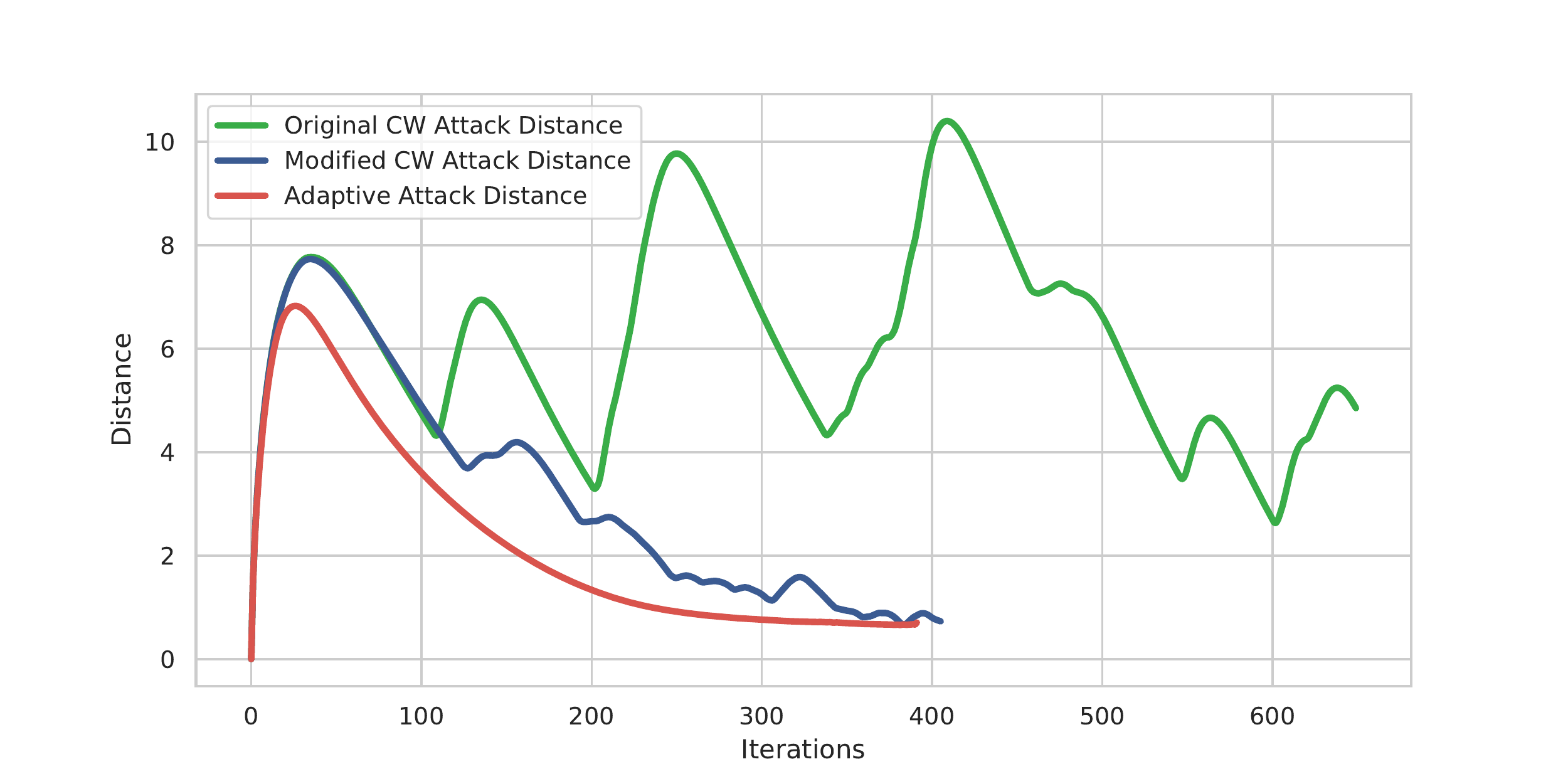}
    \caption{\textbf{$\ell_2$ distance in the Attack Process}: We attack an ImageNet example using three methods: original C\&W attack, modified C\&W attack, and our proposed Adaptive Attack and report the $\ell_2$ distance of the perturbation in the attack process. Adaptive Attack achieves a small perturbation much faster than C\&W attacks.}
    \label{fig:non-attack}
\end{figure}
\begin{figure}[!tb]
    \centering
    \includegraphics[width=0.9\linewidth]{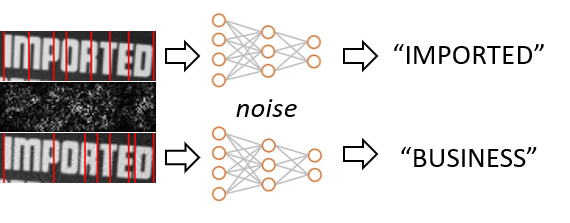}
    \caption{\textbf{An adversarial example of scene text recognition}. First row: an original image (`IMPORTED'). Second row: adversarial perturbations added to the original image. Third row: the adversarial image, incorrectly predicted as `BUSINESS'. Humans can barely recognize the difference between the two images. Red lines denote the CTC alignments.}
    \label{fig:icdar_example}
    \vspace{-0.4cm}
\end{figure}

On the other hand, accelerating adversarial attacks is a critical and non-trivial task for adversarial defenses. Adversarial Training, which trains the model with adversarial examples, has been shown as one of the most effective solutions to defend adversarial attacks~\cite{athalye2018obfuscated}. However, it is time-consuming to generate adversarial examples for training using current iterative optimization-based attacks. 
To reduce attacking time, Taesik et al. \cite{na2017cascade} trained with adversarial examples generated from C\&W attack only in the first epoch, and then adopted FGSM, a faster one-step attack but with a degraded performance on perturbations, for the rest of training. 

In this paper, we propose an \textit{Adaptive Attack} to accelerate adversarial attacking leveraging multi-task learning.
Generating adversarial examples is inherently a multi-task optimization problem, where we make an imperceptible change on the original samples while making the deep learning model to predict incorrectly. 
For classification tasks, we usually minimize distances between original examples and adversarial examples while minimizing the classification loss of adversarial examples on targeted labels.

Previous attacks simultaneously optimize two objectives, by using a weighted sum of multi-task losses. The weights of losses usually are uniformly defined~\cite{kurakin2016adversarial} or manually tuned~\cite{carlini2017towards}. For example, C\&W attack~\cite{carlini2017towards}, a widely used adversarial attack, applies a modified binary search to find a proper weight in the attack. However, the optimal weights between two tasks are strongly dependent on tasks (\eg, image distance vs. audio distance, cross-entropy loss vs. CTC loss~\cite{graves2006connectionist}). Researchers and practitioners have to carefully choose appropriate weights between task losses to achieve a good performance. Therefore, \textit{it is desirable to find a better approach to learn the optimal weights automatically}.

Multi-task learning is widely studied in many machine learning tasks, which aims to improve learning efficiency and prediction accuracy by learning multiple objectives from a shared representation~\cite{kendall2017multi}. 
Recently, Kendall et al. proposed a principled approach for multi-task weightings, by combining observation (\textit{aleatoric}) uncertainty and model (\textit{epistemic}) uncertainty. They modeled each task in a unified Bayesian deep learning framework~\cite{kendall2017multi,kendall2017uncertainties} and outperformed the separately trained models for semantic segmentation, instance segmentation, and depth regression. Their solution is limited to non-sequential learning tasks (such as image classification, image segmentation), which might not directly apply to adversarial attacks on sequential learning tasks. In Section~\ref{sequential}, we extend this idea and further derive a novel approximate solution to sequential learning tasks.

\textbf{Adversarial attack on  non-sequential learning}: 
We assume that two tasks in adversarial attacks (minimizing the classification loss and adversarial perturbations) follow probabilistic models. Our task is reformulated as attacking non-sequential classification problem by adaptively balancing tasks and tuning the weights.
We refer to our proposed method as \textit{Adaptive Attack}. Figure~\ref{fig:non-attack} illustrates that Adaptive Attack achieves a small adversarial perturbation much faster compared with strong and commonly-used attacks, C\&W attack~\cite{carlini2017adversarial} for an image recognition task. 



\textbf{Adversarial attack on sequential learning:} Current studies on adversarial examples mainly focus on non-sequential adversarial classification problems, such as image classification~\cite{szegedy2013intriguing,goodfellow2014explaining}, face recognition~\cite{sharif2016accessorize}, reinforcement learning~\cite{kos2017delving}, and semantic segmentation~\cite{hendrik2017universal}. Only very few studies target sequential learning tasks, such as speech-to-text~\cite{carlini2018audio} and reading comprehension~\cite{jia2017adversarial}, not to mention the analysis of sequential adversarial examples.  \textit{Applying Adaptive Attack to sequential learning tasks is non-trivial, due to the specific objective function and sequential properties}.

Take adversarial attacks on text recognition tasks as an example. The differences between non-sequential and sequential adversarial examples involve: i) The output of a sequential model is a varied-length label, instead of a single label. The non-sequential attacks (such as object classification model) only involve the \textit{substitution} operation (\eg, modify the original class label), while the sequential attacks consider three operations: \textit{insertion, substitution, and deletion} (\eg, insertion: coat $\rightarrow$ coat\textbf{s}, substitution: co\textbf{a}t $\rightarrow$ co\textbf{s}t, deletion: co\textbf{a}t $\rightarrow$ cot); ii) Each character in target labels needs well-aligned. The requirement of the alignment between input and output poses a challenge in generating adversarial examples; iii) Sequential models usually leverage recurrent neural networks, where the internal feature representation involves more sequential context than those in convolutional neural networks. 

Inspired by these observations, we conduct attacks on scene text recognition tasks. 
The scene text recognition is naturally a sequential learning task, which is closely related to standard classification tasks in computer vision. Measuring hidden features in the object classification task or speech recognition task is difficult, with more uncertainty on model interpretability. Adversarial attacks by modifying each character in the text image give us more intuitive explanations of how the perturbations affect the final output.

\subsection{Contributions}

We propose a novel \textit{Adaptive Attack} that directly
learns multi-task weightings without manually searching hyper-parameters, different from all previous adversarial attacks. \textit{Adaptive Attack} is a general method that can be applied to the current iterative optimization-based attacks for both non-sequential and sequential tasks. Especially for the scene text recognition attack, \textit{Adaptive Attack} accelerates adversarial attacking by three to six times. Also, we successfully attack a scene text recognition system with over 99.9\% success rate. To the best of our knowledge, it is the first work to generate adversarial examples on a scene text recognition system. 

%% file: adv.tex
In this section, we present details on our proposed \textit{Adaptive Attack} approach, which is inspired by recent findings in multi-task learning. The multi-task learning concerns the problem of optimizing a model with respect to multiple objectives~\cite{kendall2017multi}. The naive approach would be a linear combination of the losses for each task:
\begin{equation}
    \mathcal{L} = \sum_i \lambda_i \mathcal{L}_i.
\end{equation}
However, the model performance is extremely sensitive to weight selection, $\lambda_i$. It is also expensive to tune these weight hyper-parameters manually. In Bayesian modeling, we model these weight hyper-parameters using task-dependent uncertainty (homoscedastic uncertainty),
which captures the relative important confidence between tasks, reflecting the uncertainty inherent to our multiple objectives. 
\textit{Adaptive Attack} treats each task as a Gaussian distribution, where the mean is given by the model output, with an observation noise scalar $\sigma$. We will show how to relate $\sigma$ to the relative weight of each loss.
Our proposed \textit{Adaptive Attack} generates adversarial examples on both non-sequential and sequential classification tasks, which generalizes the idea of multi-task learning in~\cite{kendall2017multi}.

\subsection{Threat Model}
We assume that the adversary has access to the scene text recognition system, including the architecture and parameters of the recognition model. This type of attack is referred to as ``White-Box Attack.'' We do not consider the ``Black-Box Attack'' in this paper, which assumes the adversary has no access to
the trained neural network model. Prior work has shown that adversarial examples generated by ``White-Box Attack'' can be transferred to attack black-box
services due to the transferability of adversarial examples~\cite{papernot2016transferability}. Approximating the gradients~\cite{chen2017zoo} is another option for ``Black-Box Attack.'' 



\subsection{Basic Attack}
Give an input image $x$, the ground-truth sequential label $l=\{l_{0}, l_{1}, \cdots, l_{T}\}$,  a targeted sequential label $l'=\{l'_{0}, l'_{1}, \cdots, l'_{T'} \}$ ($l\neq l'$), and a scene text recognition model $\mathcal{F}$, generating adversarial examples can be defined as the following optimization problem:
\begin{equation}
\begin{aligned}
\label{eq:general}
&\underset{x'}{\text{minimize}} & & \mathcal{D}(x, x') \\
&s.t. & & \mathcal{F}(x') = l', \\
& & & \mathcal{F}(x) = l, \\
& & & x' \in [-1, 1]^n,
\end{aligned}
\end{equation}
where $x'$ is the modified adversarial image. $x' \in [-1, 1]^n$ ensures a valid input of $x'$. $\mathcal{D}(\cdot)$ denotes the distance between the original image and the adversarial image. 

Following C\&W Attack~\cite{carlini2017towards}, we transform the function $\mathcal{F}$ to a differentiable function, $CTC_{Loss}$. To remove the constraint of validation on new input $x'$, we introduce a new variable $w$ to replace $x'$, where $x'=\tanh(w)$. The new optimization problem is given by:
\begin{equation}
\label{eq:adaptive}
\underset{w}{\text{minimize}} \quad CTC_{Loss}(\tanh(w), l') + \lambda \mathcal{D}(x, \tanh(w)),
\end{equation}
where $CTC_{Loss}(\cdot, \cdot)$ denotes the CTC loss of the classifier $\mathcal{F}$. $\lambda$ is a task and data dependent hyper-parameter to balance the importance of being adversarial and close to the original image. People usually search for a proper $\lambda$ uniformly (log scale). In the experiments, we follow a modified binary search between $\lambda=0.01$ and $\lambda=1000$, starting from $\lambda=0.1$. For each $\lambda$, we run 2,000 iterations of gradient descent searching using Adam. We adopt an early-stop strategy to avoid unnecessary computation.

\subsection{Adaptive Attack}
It is time-consuming to search for $\lambda$ manually and find an optimal parameter $\lambda$ because $\lambda$ largely depends on individual tasks. In our experiments (Section~\ref{sec:basic_adaptive}), we compare the performance of Basic Attack with fixed $\lambda$ values, regarding the success rate of attacks, the iterations of gradient search, and the magnitude of perturbations (Figure~\ref{fig:icdar_perf}). We cannot find a proper value of fixed $\lambda$ that achieves a high success rate, small iterations, and small perturbations simultaneously. For modified binary searching of $\lambda$, as long as we conduct enough searching steps, it can always find a proper $\lambda$ to achieve a high success rate and small perturbations. However, it takes a much longer time, which makes adversarial examples hardly applicable to real-time systems.

We propose an adaptive search method. \textit{Adaptive Attack}, to generate adversarial examples. From Equation~\ref{eq:adaptive}, CTC loss $CTC_{Loss}(\cdot, \cdot)$ and Euclidean loss $\mathcal{D}(\cdot, \cdot)$ are viewed as a classification task and a regression task, respectively. Thus generating adversarial examples (Equation~\ref{eq:adaptive}) becomes solving a multi-task problem with two objectives. 

\textbf{Non-sequential Classification Tasks:} We optimize the adversarial images based on maximizing the Gaussian likelihood with uncertainty:
\begin{equation}
\label{eq:max}
\text{maximize} \quad \Pr(x'|x, l').
\end{equation}
We assume that $\Pr(x')$ and $\Pr(x, l')$ are constant values, and $x$ and $l'$ are independent variables, then
\begin{equation}
\label{eq:max2}
\begin{aligned}
    \Pr(x'|x, l') & = \frac{\Pr(x, l'|x') \Pr(x')}{\Pr(x, l')} \\ 
    & \propto \Pr(x, l'|x')  = \Pr(x |x') \Pr(l'|x')
\end{aligned}
\end{equation}
We define original input $x$ as a random variable which follows Gaussian distribution with mean $x'$ and noise scale $\lambda_1$:
\begin{equation}
\begin{aligned}
\Pr(x|x') &= \mathcal{N}(x', \lambda_1^2),\\
\log \Pr(x|x') &\propto -\frac{\|x-x'\|_2^2}{2\lambda_1^2}-\log\lambda_1^2.
\end{aligned}
\end{equation} 
For a classification task, we apply a classification likelihood to the output:
\begin{equation}
\Pr(l'|x') = Softmax(f(x')),
\end{equation}
where $f(\cdot)$ denotes the output of the neural network before softmax layer. We use a squashed version of model output with a positive scalar $\lambda_2$:
\begin{equation}
\begin{aligned}
\label{eq:class_weight}
\Pr(l'|x', \lambda_2) &= Softmax(\frac{f(x')}{\lambda_2^2}), \\
\log\Pr(l'=c|x', \lambda_2) &= \frac{f_c(x')}{\lambda_2^2} - \log\sum_{c' \neq c}\exp{\frac{f_{c'}(x')}{\lambda_2^2}}.
\end{aligned}
\end{equation}
Then we define the joint loss (optimized objective) as follows:
\begin{equation}
\resizebox{.91\linewidth}{!}{$
\begin{aligned}
\mathcal{L} =  -\log \Pr(x, l'|x') = -\log \Pr(x|x') -\log \Pr(l'|x'). \\
\end{aligned}
$}
\end{equation}
According to~\cite{kendall2017multi}, we can derive the optimization problem (Equation~\ref{eq:max},\ref{eq:max2}) by minimizing:
\begin{equation}
\mathcal{L} \propto \frac{\mathcal{L}_1(x,x')}{2\lambda_1^2} + \frac{\mathcal{L}_2(x',l')}{\lambda_2^2}+\log\lambda_1^2+\log\lambda_2^2,
\end{equation}
where $\mathcal{L}_1(x,x') = \|x-x'\|_2^2$ denoting the squared Euclidean distance and $\mathcal{L}_2(x',l')$ denotes the cross-entropy loss. We adaptively generate adversarial examples for non-sequential classification tasks without manually tuning $\lambda$. In practice, we replace $\log\lambda_i^2$ with $\eta_i$ to avoid numerical instability.

\textbf{Sequential Classification Tasks:}
\label{sequential}
Given an input sequence $X$, the network will output a probability distribution over the output domain for each frame. The probability of a given path $\pi$ can be written as:
\begin{equation}
  \Pr(\pi|x) = \prod_{t=1}^T y^t_{\pi_t}
\end{equation}
We define $\mathcal{B}$ as the conditional probability of a given labeling $l$. Let $CTC_{Loss}$ be the negative log of probabilities of all the paths given $l$:
\begin{equation}
\label{eq:ctc1}
 CTC_{Loss}(x, l) = - \log \sum_{\pi \in \mathcal{B}^{-1}(l)} \Pr(\pi| x).
\end{equation}
The $CTC_{Loss}$ of an adversarial example $x'$  with targeted label $l'$ is:
\begin{equation}
 CTC_{Loss}(x', l') = - \log \sum_{\pi \in \mathcal{B}^{-1}(l')} \Pr(\pi|x').
\end{equation}
We consider paths with probability greater than or equal to a small constant $c$: $\Pr(\pi|x')=\prod_{t=1}^T y^t_{\pi_t} \geq c$. We first assume that only two valid paths satisfy this constraint: $\pi_1, \pi_2$.
\begin{equation}
\label{eq:seq_term1}
\begin{aligned}
&-\log \Pr(l' | x') \approx -\log \left(\Pr(\pi_1|x') + \Pr(\pi_2|x') \right) \\
& \leq -\frac{1}{2}\left(\log \Pr(\pi_1|x') + \log \Pr(\pi_2|x')\right) -\log 2\\ 
& \quad \text{(Jensen's inequality)}\\
& = - \frac{1}{2} \log \prod_{t=1}^T \Pr(y^t_{{\pi_{1}}_t}) - \frac{1}{2} \log \prod_{t=1}^T \Pr(y^t_{{\pi_{2}}_t}) -\log 2 \\
& = - \frac{1}{2} \sum_{t=1}^T \left(\log \Pr(y^t_{{\pi_{1}}_t}) +\log \Pr(y^t_{{\pi_{2}}_t}) \right) -\log 2\\
& \approx \frac{1}{2} \sum_{t=1}^T \left(-\frac{A_{1,t}}{\lambda_2^2} + \log \lambda_2^2 -\frac{A_{2,t}}{\lambda_2^2} + \log \lambda_2^2\right) -\log 2 \\
& = -\frac{\sum_{t=1}^T  A_{1,t} + A_{2,t}}{2\lambda_2^2} + T \log \lambda_2^2 -\log 2,
\end{aligned}
\end{equation}
where $A_{i,t} = \log Softmax(y^t_{{\pi_{i}}_t}, f(x')), i=1,2$.

$CTC_{Loss}$ with two valid paths $\pi_1, \pi_2$ can be derived from:
\begin{equation}
\label{eq:seq_term2}
\begin{aligned}
& \quad CTC_{Loss} \approx - \log\left(\Pr(\pi_1|x') + \Pr(\pi_2|x') \right)\\
& = - \log\Pr(\pi_1|x') - \log\Pr(\pi_2|x') - \log\frac{\Pr(\pi_1|x') +\Pr(\pi_2|x') }{\Pr(\pi_1|x') \Pr(\pi_2|x') } \\
& \geq - \log\Pr(\pi_1|x') - \log\Pr(\pi_2|x') - \log\frac{2}{c}.
\end{aligned}
\end{equation}
Combining Equation~\ref{eq:seq_term1} and~\ref{eq:seq_term2}, we have an upper bound of $-\log \Pr(l' | x')$ and the joint loss $\mathcal{L}$:
\begin{equation}
\begin{aligned}
-\log \Pr(l' | x') & \leq \frac{CTC_{Loss}+\log\frac{2}{c}}{2\lambda_2^2} + T\log\lambda_2^2 -\log 2, \\
\mathcal{L} &\leq \frac{\mathcal{L}_1(x,x')}{2\lambda_1^2} + \frac{CTC_{Loss}(x',l')}{2\lambda_2^2} \\
& + \log\lambda_1^2+T\log\lambda_2^2 + \frac{1}{\lambda_2^2} -\log 2.
\end{aligned}
\end{equation}
To extend the number of valid paths from 2 to an arbitrary number $n$, the joint loss $\mathcal{L}$ satisfies:
\begin{equation}
\begin{aligned}
\mathcal{L} \leq &\frac{\mathcal{L}_1(x,x')}{2\lambda_1^2} + \frac{CTC_{Loss}(x',l')}{n\lambda_2^2}+\log\lambda_1^2+T\log\lambda_2^2 \\
& + \frac{\log n-(n-1)\log c}{n\lambda_2^2}.
\end{aligned}
\end{equation}

From our observation, the CTC loss always reduces very fast (Figure~\ref{fig:progress}). Thus we can use a small number of valid paths to generate adversarial examples. From our experiments, it works well when $n <50$. We use $n=2$ to report our results. Thus, we can generate sequential adversarial examples by minimizing the upper bound of $\mathcal{L}$:
\begin{equation}
\frac{\mathcal{L}_1(x,x')}{\lambda_1^2} + \frac{CTC_{Loss}(x',l')}{\lambda_2^2}+\log\lambda_1^2+T\log\lambda_2^2 + \frac{1}{\lambda_2^2}.
\end{equation}

%% file: analysis.tex
In this section, we first evaluate the performance of Adaptive Attack for the non-sequential classification task comparing Adaptive Attack with C\&W Attack. We then focus on the performance of Adaptive Attack for the sequential task. We attack a scene text recognition model as our use case. Besides, we investigate the generated images and the image changes during the attack for the sequential classification task.


\subsection{Non-Sequential Attack}
\label{sec:non-eval}
We evaluate non-sequential attacks for the image classification task. We use a pre-trained Inception V3 model~\cite{szegedy2016rethinking} as the victim model, which achieves 22.55\% top-1 error rate and 6.44\% top-5 error rate in the ImageNet recognition challenge. 

We compare Adaptive Attack with C\&W attack. The original C\&W attack 
achieves small perturbations but takes more iterations (See an example in Figure~\ref{fig:non-attack}). To make a fair comparison, we modify the C\&W attack by applying a more strict early-stop strategy - we stop attacking when the objective function (Equation~\ref{eq:max}) does not decrease in the past $k$ iterations. We will use the modified C\&W attack as our baseline (referred to as Basic Attack) in the paper. We set early-stop $k=20$ for Basic Attack and $k=1$ for Adaptive Attack, due to a more smooth objective function in Adaptive Attack. 

\begin{table}[!tb]
\centering
\begin{tabular}{|c|c|r|r|r|}
\hline
Methods        & Search Step & Success Rate & Distance & Iteration \\ \hline
Basic Attack & 3           & 100\%        & 1.957    & 199.298   \\ \hline
Basic Attack & 5           & 100\%        & 0.689    & 342.993   \\ \hline
Adaptive Attack      & 1           & 100\%        & 0.517    & 253.088   \\ \hline
\end{tabular}
\vspace{-0.1cm}
\caption{Performance Comparison between Basic Attack and Adaptive Attack on the ImageNet dataset.}
\label{tab:non-basic_adaptive}
\vspace{-0.3cm}
\end{table}
We evaluate Adaptive Attack on the first 2,000 images in the validation set of ImageNet. We randomly assign a new label on an image and report success if the attack generates an adversarial image predicted with this label. Table~\ref{tab:non-basic_adaptive} reports the success rate, average $\ell_2$ distance of perturbation, and average attacking iterations for modified C\&W attack (Basic Attack) and Adaptive Attack. Search step denotes the number of $\lambda$ manually searched in the Basic Attack.

Our results show that Adaptive Attack reaches small perturbations within a much less time compared to Basic Attack. 

\begin{table*}[!tb]
\centering
\small
\begin{tabular}{|c|r|r|r|r|r|r|r|r|r|}
\hline
\multirow{2}{*}{Methods} & \multicolumn{3}{c|}{IC13}                                                                         & \multicolumn{3}{c|}{SVT}                                                                           & \multicolumn{3}{c|}{IIIT5K}                                                                        \\ \cline{2-10} 
                         & \multicolumn{1}{c|}{Success Rate} & \multicolumn{1}{c|}{Distance} & \multicolumn{1}{c|}{Iteration} & \multicolumn{1}{c|}{Success Rate} & \multicolumn{1}{c|}{Distance} & \multicolumn{1}{c|}{Iteration} & \multicolumn{1}{c|}{Success Rate} & \multicolumn{1}{c|}{Distance} & \multicolumn{1}{c|}{Iteration} \\ \hline
Basic0.1                 & 99.90\%                           & 3.57                          & 1621.88                        & 99.69\%                           & 3.59                          & 1470.29                        & 99\%                              & 2.90                          & 7127.92                        \\ \hline
Basic1                   & 88.55\%                           & 1.75                          & 526.92                         & 91.55\%                           & 1.67                          & 518.99                         & 95.39\%                           & 1.77                          & 3606.26                        \\ \hline
Basic10                  & 53.38\%                           & 0.44                          & 179.75                         & 68.23\%                           & 0.47                          & 172.82                         & 39.12\%                           & 0.39                          & 1395.99                        \\ \hline
BasicBinary3             & 100.00\%                          & 1.64                          & 1531.84                        & 100.00\%                          & 1.17                          & 1442.86                        & 100.00\%                          & 2.01                          & 4097.52                        \\ \hline
BasicBinary5             & 100.00\%                          & 1.64                          & 1706.18                        & 100.00\%                          & 1.15                          & 1616.35                        & 100.00\%                          & 1.96                          & 5055.21                        \\ \hline
BasicBinary10            & 100.00\%                          & 1.58                          & 2138.86                        & 100.00\%                          & 1.11                          & 1993.47                        & 100.00\%                          & 1.94                          & 6811.86                        \\ \hline
Adaptive                 & 100.00\%                          & 2.15                          & 480.28                         & 100.00\%                          & 1.26                          & 529.90                         & 99.96\%                           & 2.68                          & 682.48                         \\ \hline

\end{tabular}

\caption{Performance Comparison between Basic Attack and Adaptive Attack on Three Scene Text Recognition Benchmarks.}
\label{tab:basic_adaptive}
\vspace{-0.3cm}
\end{table*}

\subsection{Sequential Attack}
We evaluate sequential attacks on a scene text recognition model. We compare the performance of Basic Attack and propose Adaptive Attack on three standard benchmarks. We then analyze the sequential attacks on a simulated sequential MNIST dataset.

\subsubsection{Basic Attack vs. Adaptive Attack}
\label{sec:basic_adaptive}
We conducted experiments on three standard benchmarks for cropped word image recognition: the Street View Text dataset (SVT)~\cite{wang2011end}, the ICDAR 2013 dataset (IC13)~\cite{karatzas2013icdar}, and the IIIT 5K-word dataset (IIIT5K)~\cite{MishraBMVC12}. 
We train an end-to-end deep learning model using Pytorch, based on the state-of-the-art scene text recognition approach, Convolutional Recurrent Neural Network (CRNN)~\cite{shi2017end}\footnote{we refer to the implementation on \url{github.com/bgshih/crnn} and modify the kernel size in the pooling layers for better alignment.}. 


\begin{figure}[!tb]
\centering
\includegraphics*[clip=true,width=0.85\linewidth]{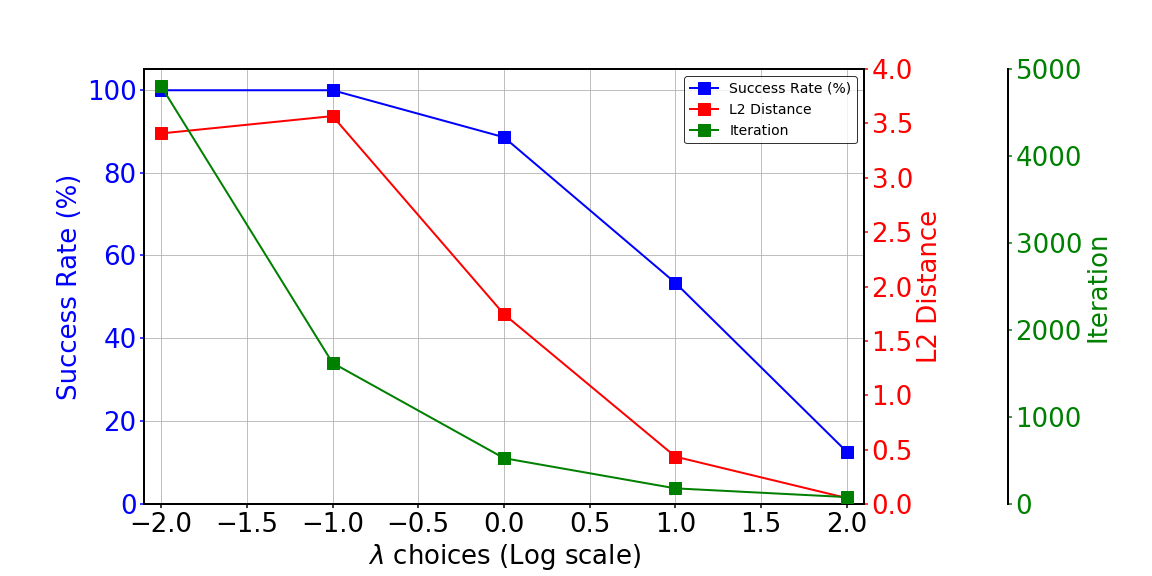}
\vspace{-0.2cm}
\caption{\textbf{Basic Attack with fixed $\lambda$ values on the \textit{IC13} dataset.} None of the $\lambda$ values can well balance CTC loss (success rate curve in blue), distance ($\ell_2$ distance curve in red), and optimizing time (iteration curve in green). }
\label{fig:icdar_perf}
\vspace{-0.3cm}
\end{figure}

\begin{figure*}[!htb]
    \centering
    \begin{subfigure}[b]{.42\columnwidth}
        \centering
        \includegraphics*[width=0.8\columnwidth]{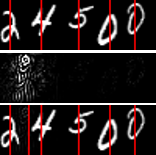}
        \caption{\centering  insertion $24500 \rightarrow 294500$}
        \label{fig:mnist_insert}
    \end{subfigure}
    \hfill
    \begin{subfigure}[b]{.42\columnwidth}
        \centering
        \includegraphics*[width=0.8\columnwidth]{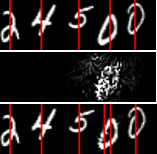}
        \caption{\centering {insertion (repeated)}  $24500 \rightarrow 245500$}
        \label{fig:mnist_insert_repeat}
    \end{subfigure}
    \hfill
    \begin{subfigure}[b]{.42\columnwidth}
        \centering
        \includegraphics*[width=0.8\columnwidth]{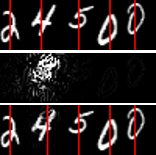}
        \caption{\centering substitution  $24500 \rightarrow 29500$}
        \label{fig:mnist_sub}
    \end{subfigure}
    \hfill
    \begin{subfigure}[b]{.42\columnwidth}
        \centering
        \includegraphics*[width=0.8\columnwidth]{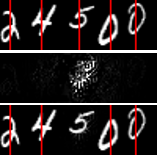}
        \caption{\centering deletion \qquad $24500 \rightarrow 2400$}
        \label{fig:mnist_delete}
    \end{subfigure}
    \vspace{-0.2cm}
    \caption{\textbf{Four types of adversarial attacks on the \textit{SeqMNIST} dataset}. The images in the first row are the same original images. After perturbing (amplified by $10\times$, second row), we generate the adversarial images (third row), which can be misclassified as different labels.}
    \label{fig:mnist_4ops}
    \vspace{-0.5cm}
\end{figure*}

We compare Basic Attack and Adaptive Attack on these benchmarks. 
The targeted sequential label is set as the common word with the same length as the original one. 
Figure~\ref{fig:icdar_perf} shows the performance of Basic Attack with fixed $\lambda$ values. We run gradient descent searches for 10,000 iterations. Early stopping is adopted to avoid unnecessary computation. We only calculate distances and iterations of the successful attacks to avoid the extremely large values when the attack fails. The results show that when we use large $\lambda$ values  (1, 10, 100), it fails to generate adversarial examples in most cases. For small $\lambda$ values (0.1, 0.01), although Basic Attack successfully generates adversarial images, it spends a much longer time and brings a larger magnitude of perturbations.

We then compare the performance of Adaptive Attack and Basic Attack using a fixed $\lambda$ or a modified binary search. Table~\ref{tab:basic_adaptive} lists our results. \textit{Basic0.1}, \textit{Basic1}, and \textit{Basic10} denote Basic Attack with fixed $\lambda$ values: 0.1, 1, and 10, respectively. \textit{BasicBinary3}, \textit{BasicBinary5}, \textit{BasicBinary10} denote Basic Attack with 3, 5, and 10 steps of binary searching. We set the initial $\lambda$ as 0.1, which is the best $\lambda$ according to the results of fixed $\lambda$ values (Figure~\ref{fig:icdar_perf}).

From Table~\ref{tab:basic_adaptive}, we observe that both Adaptive Attack and Basic Attack with a modified binary search can successfully generate adversarial examples on the scene text recognition model. Basic Attack with fixed $\lambda$ values cannot achieve both success rate and low distance of perturbations. Adaptive Attack conducts attacks much faster ($3\sim6\times$) than Basic Attack with a modified binary search. Although Basic Attack achieves smaller perturbations than Adaptive Attack, it is reasonable for binary search method to have a finer tuning on the $\lambda$ if the initial $\lambda$ value is properly set and the number of binary search iterations is large enough.

\subsection{Analysis of Sequential Attack}
\label{sec:sub_analysis}

To dig into the phenomenon of sequential adversarial examples, we generate adversarial examples on a simple sequential classification task. We first simulate a sequential digit dataset by concatenating digit images in the MNIST dataset and fit them into a 32x100 pixel box. The training and test sequential digits are generated from MNIST training and test sets. We refer to this dataset as \textit{SeqMNIST}. The first row in Figure~\ref{fig:mnist_4ops} illustrates an example (`24500'). We then trained our model with \textit{SeqMNIST}. 

We analyze three types of common adversarial operations on targeted sequential labels: insertion, substitution, and deletion. We perform these operations on one digit and remain the rest unchanged. We also include another operation that inserts a repeated digit (\eg, `24500' $\rightarrow$ `245500'). We perform an adversarial attack on the \textit{SeqMNIST} dataset and exam 100 adversarial images for each operation.

\label{sec:iteration}
\begin{figure}[!tb]
    \centering
    \includegraphics*[width=0.8\linewidth]{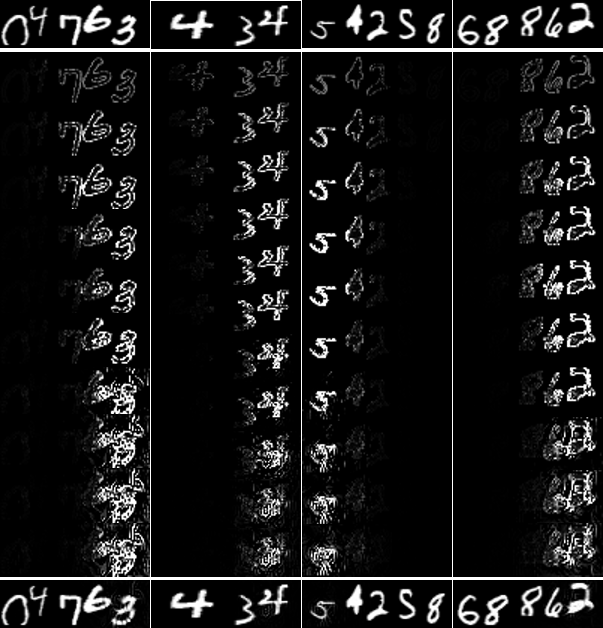}
    \caption{\textbf{Attack Process.} We demonstrate four adversarial examples on the \textit{SeqMNIST} dataset, including insertion (`04763' $\rightarrow$ `047673'), insertion repeated (`434' $\rightarrow$ `4334'), substitution (`54258' $\rightarrow$ `94258'), and deletion (`68862' $\rightarrow$ `6886'). First row: original images. Last row: adversarial images. The middle rows: adversarial perturbations (amplified by $10\times$) added on the original images in the order of iterations.}
    \label{fig:progress}
    \vspace{-0.2cm}
\end{figure}

\textbf{CTC alignments in the adversarial examples:} We observe that most CTC alignments are stable against adversarial examples (Figure~\ref{fig:mnist_4ops}). When added perturbations on the images, only the CTC alignments surrounding the targeted labels will be changed. We investigate the four operations: (1) insertion, (2) repeated insertion, (3) substitution, and (4) deletion.

\textbf{Visualizing the attack process:} We analyze the process of generating adversarial examples and focus on the change of the CTC alignment and the perturbation distance. 
Figure~\ref{fig:progress} visualizes the attack process of a \textit{SeqMNIST} sample.
The adversarial perturbations are firstly added to the edges of digits surrounding the targeted positions, 
(the new positions based on various adversarial operations) 
where the gradients are larger than the others. The CTC alignments will be aligned to the targeted positions very quickly (in most cases, less than 100 iterations). After the CTC alignments get stable, the alignments will not be changed. The later changes mainly minimize the adversarial perturbations while keeping the prediction unchanged. In a few iterations, the adversarial perturbations look similar to the targeted digits. In the final iteration, it will be minimized to an `abstract' version of targeted digits, which can not be recognized by humans. Based on previous observations, CTC alignments only change in the first few iterations and appear close to the targeted (inserted/ substituted/ deleted) positions. After that, the adversarial attack will focus on minimizing the magnitude of perturbations.

%% file: conclusion.tex
We have proposed a novel approach to learn multi-task weights without manually tuning the hyper-parameters. The proposed Adaptive Attack method substantially speeds up the process of adversarial attacks for both non-sequential and sequential tasks.
Our proposed attacks successfully fooled the scene text recognition system with over $99.9\%$ success rate on three standard benchmark datasets.
We will investigate defense mechanisms against \textit{Adaptive Attack} for both non-sequential and sequential tasks in our future work.